\documentclass[10pt,twocolumn,letterpaper]{article}

\usepackage[pagenumbers]{cvpr} % To force page numbers, e.g. for an arXiv version

\definecolor{cvprblue}{rgb}{0.21,0.49,0.74}
\usepackage[pagebackref,breaklinks,colorlinks,allcolors=cvprblue]{hyperref}
\usepackage{booktabs}   
\usepackage{graphicx} 
\usepackage{multirow}    
\usepackage{makecell}  
\usepackage{amsmath}   
\usepackage{booktabs}
\usepackage{subcaption}
\usepackage{float}
\usepackage{placeins}
\newcommand{\et}[2]{${#1}^{\pm{#2}}$}
\newcommand{\etb}[2]{$\mathbf{{#1}}^{\pm{#2}}$}
\newcommand{\ets}[2]{$\underline{{#1}}^{\pm{#2}}$}

\newcommand{\name}{RMD\xspace}

\title{RMD: A Simple Baseline for More General Human Motion Generation \\ via Training-free Retrieval-Augmented Motion Diffuse}

\author{
Zhouyingcheng Liao$^{1}$ , 
Mingyuan Zhang$^{2}$, 
Wenjia Wang$^{1}$,
Lei Yang$^{3}$\footnotemark[2]\,,
Taku Komura$^{1}$
\\
$^{1}$The University of Hong Kong  \quad	
$^{2}$Nanyang Technological University \quad
$^{3}$SenseTime Research\\
{\small \url{https://zycliao.github.io/rmd}}
}

\begin{document}
% \maketitle

\twocolumn[{
    \renewcommand\twocolumn[1][]{#1}
    \maketitle

    \vspace{-25pt}
    \begin{center}
        \includegraphics[width=0.95\textwidth]{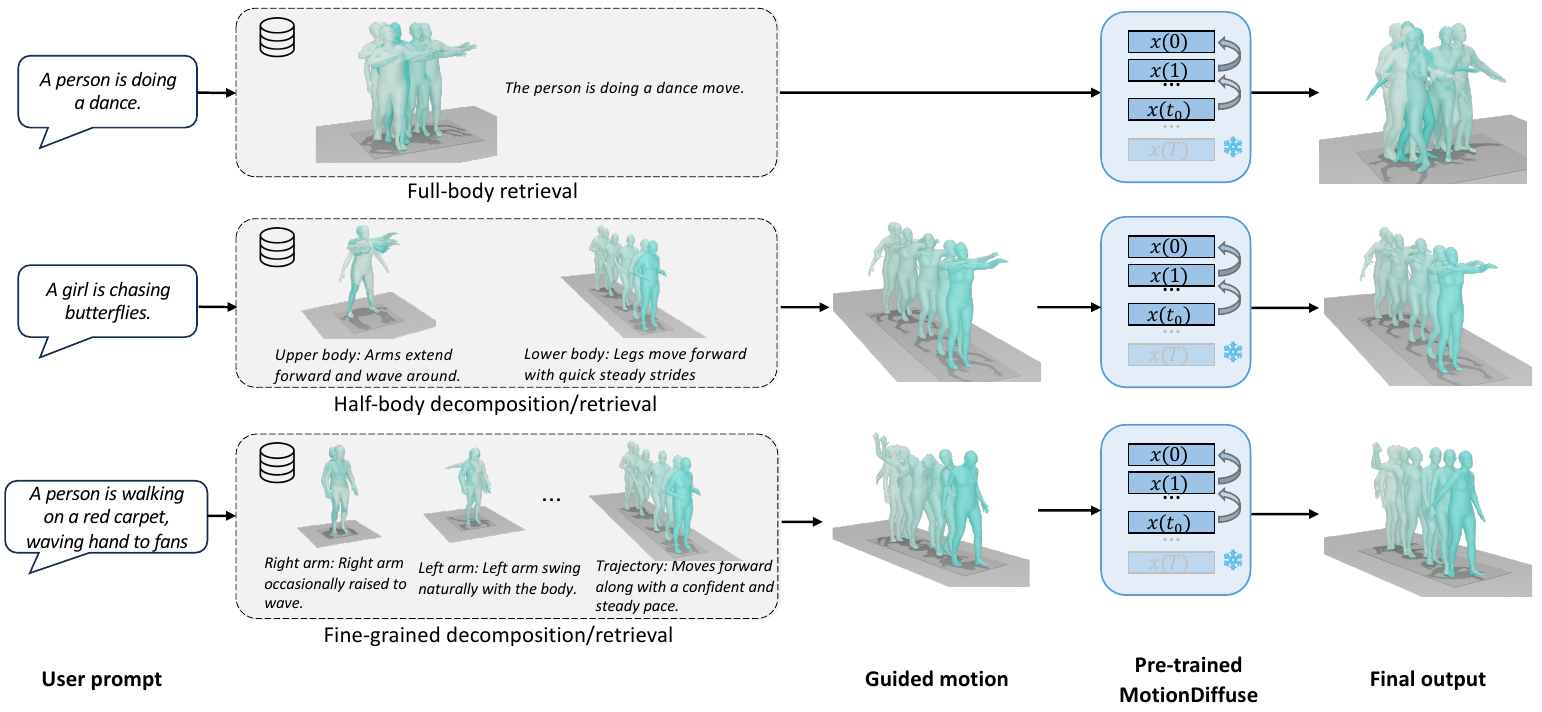}
        \captionof{figure}{
        Existing methods struggle with out-of-distribution motion generation due to two main challenges: (1) The compositional complexity of human motion makes it difficult for training sets to cover all possible full-body motions; (2) Diverse motion descriptions create a persistent gap between testing and training prompts.
        We propose a simple baseline \textbf{\name}, a retrieval-augmented, training-free method with two stages: \textbf{motion retrieval}, using a decompose-retrieve-recompose hierarchical strategy of 3 different levels to bridge the aforementioned gap, and  \textbf{motion diffusion}, refining the composed motion with a pre-trained diffusion model to enhance body coordination and enrich generation diversity.}
    \label{fig:teaser}
    \end{center}
}]
\renewcommand{\thefootnote}{\fnsymbol{footnote}}
\footnotetext[2]{Corresponding author.}   
\begin{abstract}
While motion generation has made substantial progress, its practical application remains constrained by dataset diversity and scale, limiting its ability to handle out-of-distribution scenarios.
To address this, we propose a simple and effective baseline, RMD, which enhances the generalization of motion generation through retrieval-augmented techniques.
Unlike previous retrieval-based methods, RMD requires no additional training and offers three key advantages:
(1) the external retrieval database can be flexibly replaced;
(2) a hierarchical retrieval module to reuse body parts from the motion database, with an LLM~(large Language Model) facilitating splitting and recombination; and
(3) use a pre-trained motion diffusion model as a prior to improve the quality of motions obtained through retrieval and direct combination.
Without any training, RMD achieves state-of-the-art performance, with notable advantages on out-of-distribution data.
\end{abstract}    

\section{Introduction}
\label{sec:intro}

% \begin{figure*}
%     \centering
%     \includegraphics[width=\linewidth]{figs/teaser.png}
%     \caption{Existing methods struggle with out-of-distribution motion generation due to two main challenges: (1) Human motion is inherently compositional, making it exponentially difficult for a training set to cover all possible full-body motions; (2) The diversity of motion descriptions creates a persistent gap between testing and training prompts.
%     To address this, we propose \name, a retrieval-augmented, training-free baseline with two stages, namely motion retrieval and motion diffusion. The former stage employs a decompose-retrieve-recompose strategy to bridge the aforementioned gap, while the latter stage adopts a pre-trained diffusion model to refine the composed motion, enhancing coordination across body movements.}
%     \label{fig:teaser}
% \end{figure*}

%% Our goal
Motion generation has been widely used in film and gaming industries. Although text-to-motion~\cite{zhang2022motiondiffuse,tevet2023human,zhang2023remodiffuse,zhang2023finemogen,qing2023storytomotion,zhang2024large,guo2024momask} has made significant progress, it still performs poorly on out-of-distribution (OOD) text inputs in real-world applications.

%% Current Method
Current methods can be divided into two primary categories.
The first category of approaches \cite{zhang2022motiondiffuse, tevet2023human, zhang2023finemogen, zhang2024large, guo2024momask, emdm, tlcontrol} directly generates motion from text. Unlike tasks such as text or image generation, which benefit from vast training datasets, 3D human motion datasets are limited in diversity and scale. This limitation poses challenges for handling out-of-distribution (OOD) cases in motion generation.
The second category \cite{petrovich2023tmr, bensabath2024cross, zhang2023remodiffuse, cai2024digital} retrieves motions based on the given text and incorporates them into the generation model. This approach enhances performance on specific datasets. However, they require additional training and still struggle to deliver satisfactory results for OOD scenarios.

%% High-level idea
To address gaps in motion data, human expertise is sometimes required. For instance, if the motion database lacks a motion of simultaneously walking and waving, a human artist might blend lower-body walking with upper-body waving, making minor adjustments to create the desired motion.
Inspired by artists' effective approach to creating an unseen motion, we developed a simple, training-free baseline called \name (\textbf{R}etrieval-augmented \textbf{M}otion \textbf{D}iffuse) to enhance generalized motion generation.

%% Our Method
\name consists of two main stages, namely motion retrieval stage and motion diffusion stage.
In the motion retrieval stage, relevant motions are selected and combined based on user input text prompt, using a customizable external motion database.
In the motion diffusion stage, the combined motions are refined by a pre-trained motion diffusion model to enhance motion quality.

To improve generalization and flexibility, we employ an external motion database for retrieval. The key challenge lies in locating the appropriate full-body or body-part-specific motions.
Given the multi-granular nature of motions, we designed a simple multi-level retrieval pipeline using an LLM.
When a text input is received, the pipeline first assesses if the motion could be split by body part.
For example, “backflip” is a complete motion, while “walking and waving” is a composite. The former is searched as a whole in the database, whereas the latter is divided into upper and lower body components for separate retrieval.
For input requiring finer details, the LLM further decomposes the motion into specific body parts and retrieves each part individually.
Examples can be seen in Figure~\ref{fig:teaser}.
When the multiple body part motions are retrieved, they are assembled according to the LLM’s decomposition.
Apparently, direct combinations of different body parts can lead to misalignment issues since each retrieved part is not aware of the overall posture and semantic coherence.
To address this, we use a pretrained motion diffusion model as a prior to improve motion quality and diversity.
Inspired by SDEdit~\cite{meng2021sdedit}, we adopt a noise-and-denoise scheme.
In the diffusion stage, \name first adds noise to the combined motion from the retrieval stage and then leverages a trained motion diffusion model to denoise it under the guidance of input text, refining the motion for better quality.
%
% In our experiments, we use MotionDiffuse for its simplicity and effective training performance.

%% Experiments
Experiments demonstrate that \name achieves state-of-the-art performance with a simple, training-free pipeline, excelling particularly on out-of-distribution data.
On the standard benchmark HumanML3D~\cite{guo2022generating}, using the training set as the motion retrieval database still yielded performance gains, suggesting that current training algorithms do not yet fully exploit the knowledge in training datasets.
In addition to standard in-domain testing, we use Mixamo dataset~\cite{mixamo} to evaluate various algorithms under \emph{cross-domain} scenario.
Results show that our simple retrieval-based design can outperform existing approaches. Existing algorithms exhibit low overall accuracy in cross-domain settings, highlighting significant room for improvement in generalization. Furthermore, methods that perform well on HumanML3D do not necessarily generalize to Mixamo, indicating potential overfitting in current approaches.
We also designed practical prompts based on popular video content and conducted a user study, which revealed that, while accuracy on HumanML3D saturates, visualizations still fall short of user expectations, leaving a gap before achieving the ease of use seen in text or image generation applications.

%% Summary
In summary, we present a lightweight, easy-to-produce, and high-performance method, systematically benchmarking existing methods under OOD scenarios and setting a new baseline for future general motion generation.
\section{Related Work}
\label{sec:related}
\paragraph{Text-driven motion generation.}
%
% Recent advancements in human motion generation have notably shifted towards integrating diverse methodologies to enhance both the realism and versatility of generated motions.
% %
% Traditional approaches like JL2P and MotionCLIP have focused on creating a unified embedding space for text and motion, aiming to bridge the gap between linguistic descriptions and physical movements. These methods have been foundational, but recent trends lean towards employing more sophisticated models like variational autoencoders and diffusion models to increase the diversity and controllability of motion sequences. For instance, TEMOS and T2M-GPT utilize transformer architectures and variational techniques to improve upon deterministic models, allowing for more nuanced and varied motion outputs.
% Moreover, the advent of diffusion models, exemplified by MotionDiffuse and MDM, has introduced a new paradigm in motion synthesis, offering capabilities for multi-level manipulation and geometric constraint handling. Models like MotionGPT and MotionLLM are pushing towards a holistic integration of language and motion, incorporating large language models' world knowledge to not only generate but also edit and understand motions in a more contextually rich manner. These models aim to perform a variety of motion-related tasks, from generation to fine-grained editing, by leveraging the expansive capabilities of large-scale, multi-modal learning frameworks.

% TODO: finemogen, momask

Text-to-motion is a prominent topic in conditional motion generation, requiring models to interpret text and generate corresponding motions. 
The advent of diffusion models, such as MotionDiffuse~\cite{zhang2022motiondiffuse} and MDM~\cite{tevet2023human}, has introduced a new paradigm in motion synthesis.
Models such as MotionGPT~\cite{zhang2023motiongpt} and MotionLLM~\cite{chen2024motionllm} integrate language and motion by leveraging Large Language Models (LLM), enabling contextually rich motion generation, editing, and understanding.
Story-to-motion~\cite{qing2023storytomotion} uses LLMs for semantic control and enhances transitions with transformers. 
FineMoGen~\cite{zhang2023finemogen} improves the generation of complex, temporally coordinated motions, by explicitly modeling spatio-temporal composition constraints.
STMC~\cite{petrovich2024multi} proposes a test-time denoising method to allow users to specify a multi-track timeline of several prompts for different body parts.
MoMask~\cite{guo2024momask} incorporates masked modeling for text-to-motion, representing human motion as multi-layer discrete motion tokens.
LMM~\cite{zhang2024large} unifies mainstream motion tasks into a single generalist model.
These models leverage multi-modal learning for diverse motion tasks, but their generalization to real-world out-of-distribution scenarios remains limited.

% \paragraph{Retrieval Augmented Generation.}
\paragraph{Retrieval-augmented text-to-motion.}
% motion matching,
% CLIP,
% remodiffuse, tmr, tmr++, DLP(MoMat-MoGen)
% RAG LLM? cite a survey

It has been proven that Retrieval-Augmented Generation (RAG) is effective in enhancing generative models for LLMs~\cite{lewis2020retrieval,guu2020retrieval,ram2023context,gao2023retrieval}, image generation~\cite{chen2022re,sheynin2022knn,blattmann2022retrieval}, and video generation~\cite{he2023animate}, and some work has also incorporated retrieval into text-to-motion.
TMR~\cite{petrovich2023tmr}, built on TEMOS~\cite{petrovich2022temos}, uses contrastive training like CLIP to learn a cross-modal embedding space for join motion retrieval and synthesis.
TMR++~\cite{bensabath2024cross} extends TMR by appling LLM to augment text in the retrieval pipeline, partially addressing dataset biases in standard text-motion benchmarks and narrowing the domain gap.
ReMoDiffuse~\cite{zhang2023remodiffuse} incorporates hybrid retrieval to improve denoising in MotionDiffuse.
DLP~\cite{cai2024digital} introduces the MoMat-MoGen algorithm for human-human interactions, integrating motion matching for quality and motion generation for diversity.
However, these methods rely on training data distributions, limiting their ability to handle out-of-distribution scenarios effectively.

% \paragraph{Diffusion-based Editing.}
\paragraph{Diffusion model as a prior.}

% SDEdit + diffusion inversion?

Pretrained diffusion models have been proved to be strong priors for various downstream tasks, such as image editing~\cite{huang2024diffusion}, video editing~\cite{sun2024diffusion,ouyang2024i2vedit} and motion composition~\cite{shafir2023human}.
SDEdit~\cite{meng2021sdedit} first adds noise to a manipulated input image and then denoises it using the diffusion model prior, enhancing both the realism and generalization.
Delta Denoising Score (DDS)~\cite{hertz2023delta} is a diffusion-based technique that uses the Score Distillation Sampling (SDS)~\cite{poole2022dreamfusion} score to optimize and guide an image toward a text-specified target, leveraging the generative prior of text-to-image diffusion models.
PriorMDM~\cite{shafir2023human} uses a pretrained MDM as a prior for sequential composition, demonstrating the potential of using priors at inference to refine temporal transitions.
Diffusion Noise Optimization (DNO)~\cite{karunratanakul2024optimizing} optimizes latent noise in pre-trained motion diffusion models to enhance various motion tasks without retraining.
Inspired by SDEdit's simple and effective approach, we employ such noise-and-denoise mechanism with a pre-trained diffusion model as a prior, to refine motions composed from retrieval.
\newcommand{\tx}{\text{text}}
\newcommand{\TX}{\text{TEXT}}
\newcommand{\motion}{\textbf{x}}
\newcommand{\gmotion}{\textbf{x}^{(g)}}

\section{Method}
\label{sec:method}

\begin{figure*}
    \centering
    \includegraphics[width=0.94\linewidth]{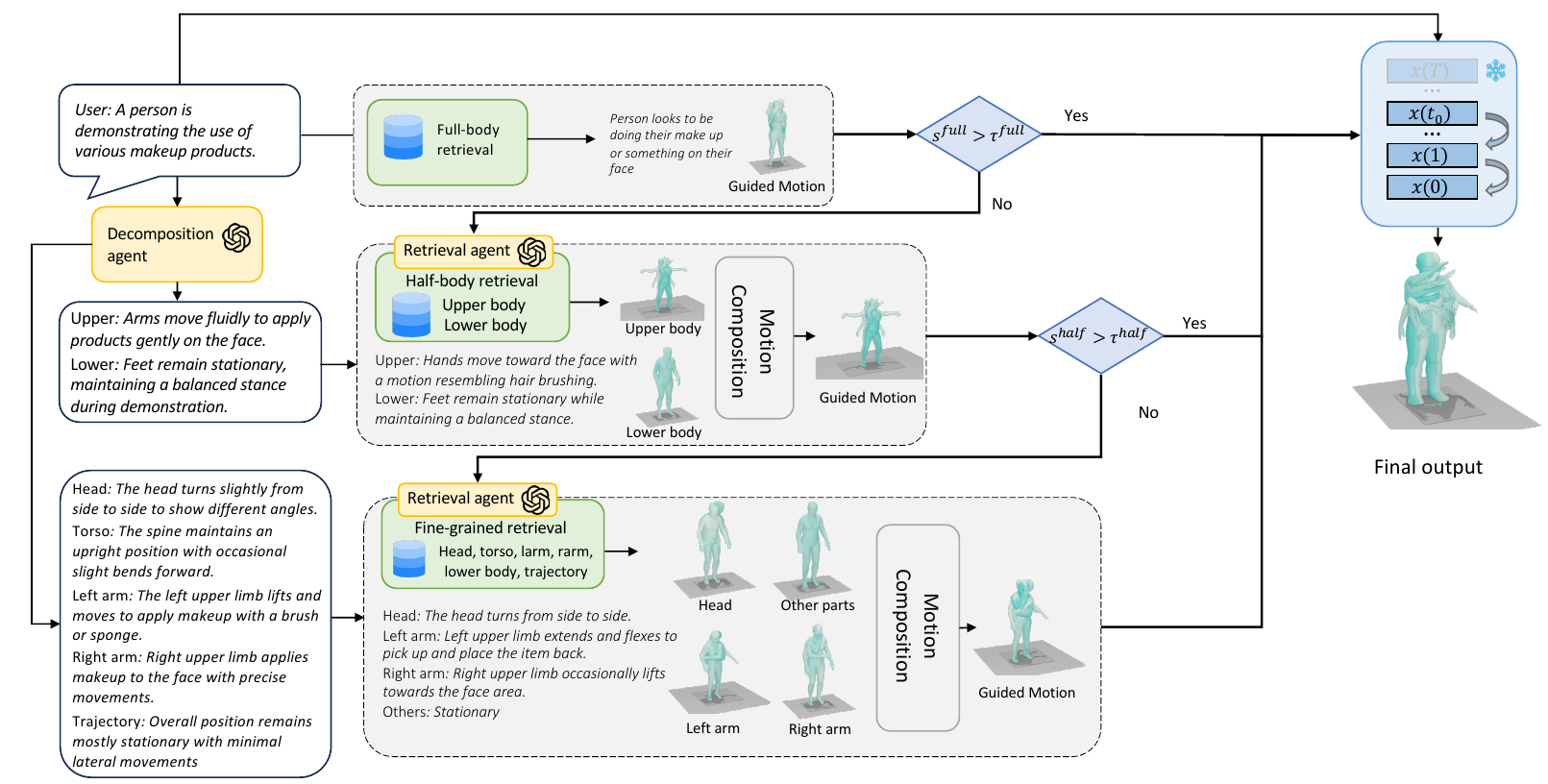}
    \caption{\textbf{Method overview of RMD.}
    Given a query text prompt, RMD uses a \emph{Decomposition Agent} to split the prompt into body parts descriptions and a \emph{Retrieval Agent} to search for corresponding motions.
    In the first stage, a hierarchical retrieval strategy is employed, prioritizing full-body to fine-grained motions. The process stops once the retrieval score meets the threshold, and the retrieved body parts are recomposed into a full motion, serving as the guided motion.
    In the second stage, RMD leverages a pre-trained motion diffusion model to refine the guided motion with the original query prompt, yielding the final motion.}
    \label{fig:method}
\end{figure*}

% \subsection{Pipeline Overview}
Fig.~\ref{fig:method} illustrates our framework.
First, we construct a motion database $\mathcal{M} = \{ (\motion_i, \text{text}_i)\}$, where each motion description $\text{text}_i$ is decomposed into half-body descriptions and fine-grained descriptions.
Given a text prompt $\text{text}_p$ describing a human motion, we employ a hierarchical strategy to determine whether the prompt should be used directly, decomposed into half-body motions, or further split into fine-grained motions.
After retrieving motions from each body part, we recompose them to form a full-body motion, referred to as the guided motion.
However, the composition might bring extra artifacts.
Thus, we apply SDEdit with a pre-trained diffusion model, using original prompt $\text{text}_p$ to refine the guided motion for an optimal balance of semantic accuracy and motion quality.

\subsection{LLM-based Motion Decomposition Agent}
% Although existing motion generation methods produce good results on standard benchmarks such as HumanML3D, they fail to produce convincing results for out-of-distribution (OOD) texts.
%
% While motion retrieval provides more clue to the generation model, the size of the retrieval database usually is still limited and cannot handle most OOD cases.
%
Human motion is inherently compositional, \ie, it is composed of distinct movement of various body parts, making it hard to a motion database containing all kinds of human movement.
Our key insight is that by decomposing a full-body motion into more atomic sub-motions of different body parts and recomposing them, existing motion databases can cover a much wider variety of OOD motions.
Meanwhile, for the same motion, there exist numerous descriptions of 
different phrasing and level of details.
%different granularity and phrasing.
%
Even for a motion seen during training, the model could still fail to generate if the test prompt diverges from the training text distribution.
Decomposing the training motion texts and the test prompt using the same LLM agent helps bridge this gap.
\par
For a prompt describing a full-body motion $\tx^{full}$, we use an LLM agent to decompose it into descriptions of two levels of granularity: \\
1) \textbf{Half-body level}, decomposing the motion into upper body and lower body $\TX^{half} = \{ \tx^{upper}, \tx^{lower} \}$; \\
% 2) \textbf{Fine-grained level}, producing sub-motions of head, torso, left arm, right arm, lower body, and global trajectory,
2) \textbf{Fine-grained level}, decomposing the full-body motion into sub-motions for six body parts\footnote{head, torso, left arm, right arm, lower body, and global trajectory},
$\TX^{fine} = \{ \tx^{head}, \tx^{torso}, \tx^{larm}, \tx^{rarm}, \tx^{lower}, \tx^{traj}\}$.
% lei: change torso to torso if only used here
%
\par
For motions with multiple text prompts, we concatenate them and inform the LLM that they describe the same motion.
%
% We empirically found that this strategy produces better results than treating them as different items and decomposing them separately.
Empirically, this approach yields better results than treating the prompts as separate items and decomposing them individually.

%%%%%%%%%%%%%%%%%%%%%%%%%%%%%%%%%%%%%%%%%%%%%%%%%%%%%%%%%%%
\subsection{Hierarchical Motion Retrieval}
To perform retrieval, we start by defining a feature set for each data entry: $\{ \tx, l \}$, where $l$ is the length of the motion, and $\tx \in \{ \tx^{full}, \TX^{half}, \TX^{fine} \} $. 

We then collect all the training motions to construct the retrieval database.
For a given query prompt, we compute its corresponding feature set and search for its best match in the database.
The retrieval for each body part is handled independently.

\paragraph{Naive retrieval.}
Our approach extracts features to measure similarities between input text descriptions and database entities. 
Using the pre-trained CLIP model~\cite{radford2021learning}, we generate text embeddings for both queries and data points.
For each data entry $(\motion_i, \text{text}_i)$, we derive the text-query feature $f^t_i = E_T(\text{text}_i)$ using CLIP's text encoder $E_T$.
Following ReMoDiffuse~\cite{zhang2023remodiffuse}, we also consider motion sequence length as a crucial feature for retrieval to compute the similarity score $s_i$:
\begin{equation}
\begin{aligned}
&s_i = \langle f^t_i, f^t_p \rangle \cdot e^{-\lambda \cdot \gamma}, \quad \gamma = \frac{| l_i - l_p |}{\max\{l_i, l_p\}},
\label{eq:score}
\end{aligned}
\end{equation}
where $\langle \cdot, \cdot \rangle$ represents cosine similarity and $l_i$ denotes the length of motion sequence $\motion_i$, and $p$ denotes the query data. The score increases with higher semantic alignment and closer length matching, with $\lambda$ controlling the length difference's impact.

\paragraph{LLM-based retrieval agent.}
Although CLIP provides a simple solution to matching the semantic information, it is not robust enough as one motion can be described by sentences phrased very differently.
To enhance retrieval robustness, we propose to leverage the world knowledge of LLM .
%%%
For each query prompt $\tx^{full}_i$, we perform LLM decomposition $k$ times to obtain $\{\TX^{half}_i, \TX^{fine}_i | i \in \{1,...,k\}\}$.
Then, each resulting description undergoes naive retrieval to find its best match in the database.
For each body part, we obtain its $k$ candidates and the corresponding textual descriptions.
%
% We then provide the LLM with these $k$ descriptions and ask which matches the original query prompt the best.
%
% We select the highest score out of the $k$ retrieval as the similarity score $s_i$.
%
Given these $k$ descriptions, we then prompt the LLM to select the description that best aligns with the original query prompt, taking the highest match score as the similarity score $s_i$.
%
% In practice, we empirically chose $k=5$.
Empirically, $k=5$ yields a good balance between accuracy and efficiency.
%%%
An alternative strategy is to perform decomposition only once, retrieve top-$k$ matches, and prompt the LLM to choose among these $k$ candidates.
However, we found the top-$k$ matches usually have very similar phrasing.
In the case where the best match has a different phrasing from the query prompt, this strategy might fail to retrieve the best match.

\paragraph{Hierarchical retrieval.}
Decomposing full body motion into distinct body parts allows for greater flexibility in composing OOD motions through retrieval.
However, composing motions from different sources can result in unnatural movements.
Therefore, we prioritize full-body retrieval or decomposition into fewer body parts whenever possible.
To determine the decomposition approach, we first evaluate if the similarity score for full-body retrieval $s^{full}_i$ exceeds a specified threshold $\tau_{full}$.
If it does, we proceed with full-body retrieval.
If not, we check whether the average similarity score for half-body decomposition $(s^{upper}_i + s^{lower}_i) / 2$ is greater than the threshold $\tau_{half}$.
If this criterion is met, the half-body decomposition is used.
Otherwise, the fine-grained decomposition is applied.

\paragraph{Motion Composition.}
In the case where motion decomposition is chosen instead of direct full-body retrieval, we need to re-compose all the body parts to form a full-body motion.
We select the quaternion of all corresponding joints for each body part and re-combine them.
If the half-body decomposition is chosen, we copy the global translation from the retrieved lower-body.
Since the retrieved motions might have different lengths from the query motion length, we rescale the retrieved quaternion and translation using SLERP and linear interpolation to fit the query length.
Then, we convert the joint quaternion and global translation into the pose representation by Guo \etal~\cite{guo2022generating}: $(r^{va},r^{vx}, r^{vz},r^h, \mathbf{j}^p, \mathbf{j}^v, \mathbf{j}^r)$. The root motion is described by four scalars: angular velocity around the Y-axis ($r^{va}$), linear velocities along X and Z axes ($r^{vx}, r^{vz}$), and height ($r^h$). Joint information is encoded in three matrices: local positions ($\mathbf{j}^p \in \mathbb{R}^{J \times 3}$), velocities ($\mathbf{j}^v \in \mathbb{R}^{J \times 3}$), and 6D continuous rotations ($\mathbf{j}^r \in \mathbb{R}^{J \times 6}$), where $J$ is the number of joints. 

%%%%%%%%%%%%%%%%%%%%%%%%%%%%%%%%%%%%%%%%%%%%%%%%%%%%%%%%%%%
\subsection{Retrieval-Augmented Motion Diffusion}
Through retrieval and composition, we obtain a full-body motion $\gmotion$ that already roughly matches the input prompt.
However, since the motions of different body parts are retrieved from different sources, composing them together might bring extra unnaturalness.
For example, the motions of the upper body and the lower body might have different movement ranges and do not coordinate well.
%
% To solve that, we utilize the powerful motion prior of a pre-trained diffusion model to refine the guided motion using the manner of SDEdit~\cite{meng2021sdedit}.
Previous works~\cite{huang2024diffusion,sun2024diffusion,shafir2023human} have shown the potential of pre-trained diffusion models as priors across various tasks. Although many methods are intricately designed, for simplicity and reproducibility, we refine the guided motion following the SDEdit~\cite{meng2021sdedit} approach, given its widespread use.
%We also compared additional alternative methods in our experiments.
%
\par
SDEdit leverages a key insight about reverse Stochastic Differential Equations (SDEs): they can be solved starting from any intermediate time $t_0$ between 0 and 1, not just from $t_0 = 1$ as in previous SDE-based generative models.
More specifically, we first choose a starting time $t_0$ between 0 and 1.
%
% Then a Gaussian noise is added to the guide: $\mathbf{x}(t_0) \sim \mathcal{N}(\motion; \sigma^2(t_0) \mathbf{I})$.
Then we initialize SDE with a noisy input $\motion$ by adding Gaussian noise to the guided motion $\gmotion$: $\motion = \gmotion + \sigma(t_0) \mathbf{z}$, where $\mathbf{z} \sim \mathcal{N}(\mathbf{0};\mathbf{I})$.
%
% We use DDIM to solve the reverse SDE from time $t=t_0$ to $t=0$ to generate the final motion $\motion(0)$
We use DDIM to solve the reverse SDE from time $t=t_0$ to $t=0$ to progressively remove the noise to obtain denoised final motion $\motion(0)$.
% We denote this complete process as
% \begin{equation}
%  \mathrm{SDEdit}(\gmotion; t_0, \theta) .    
% \end{equation}
We denote this complete process as $\mathrm{SDEdit}(\gmotion; t_0, \theta)$. Given a total denoising step $N$ and the noised $\motion$ as input, for a denosing step $n$, $\mathrm{SDEdit}$ first samples $\mathbf{z} \sim \mathcal{N}(\mathbf{0};\mathbf{I})$, then it updates $\motion$ as:
\begin{equation}
\begin{split}
 \epsilon = \sqrt{\sigma^2(t) - \sigma^2(t - \Delta t)} \\
  \motion = \motion + \epsilon^2 \mathbf{s_{\theta}} (\motion, t) + \epsilon\mathbf{z},
\end{split}
\end{equation}
where $t = t_0 \frac{n}{N}$, $\Delta t = \frac{t_0}{N}$, $\mathbf{s_{\theta}} (\motion, t)$ denotes the pre-trained score model.

The choice of $t_0$ (alongside the discretization steps used by the SDE solver) is the key hyperparameter in SDEdit.
It provides the user the flexibility to balance between the retrieved and composited motion and the diffusion prior. 
When $t_0$ is small, it maintains similarity to the guide $\gmotion$.
As it $t_0$ increases, the generated motion becomes closer to the diffusion prior.
When $t_0=1$, the process is a standard diffusion sampling without the motion guide.
\section{Experiment}

\begin{table*}[th]
    \centering
    \scalebox{0.85}{
    \begin{tabular}{l l c c c c c c c}
    \toprule
    % \multirow{2}{*}{Methods} & \multicolumn{4}{c}{\DN (Coarse-grained)} & & \multicolumn{4}{c}{HumanAct(Fine-grained)} \\
    % \cline{2-5}
    % \cline{7-10}
    %                 & FID$\downarrow$ & Accuracy$\uparrow$ & Diversity$\rightarrow$& MModality$\rightarrow$ &  & FID$\downarrow$ & Accuracy$\uparrow$ & Diversity$\rightarrow$ & MModality$\rightarrow$\\
    \multirow{2}{*}{Datasets} & \multirow{2}{*}{Methods}  & \multicolumn{3}{c}{R Precision$\uparrow$} & \multirow{2}{*}{FID$\downarrow$} & \multirow{2}{*}{MM Dist$\downarrow$} & \multirow{2}{*}{Diversity$\rightarrow$} & \multirow{2}{*}{MultiModality$\uparrow$}\\

    \cline{3-5}
       ~& ~ & Top 1 & Top 2 & Top 3 \\
    %\midrule
     %   \textbf{Real motions} & \et{0.511}{.003} & \et{0.703}{.003} & \et{0.797}{.002} & \et{0.002}{.000} & \et{2.974}{.008} & \et{9.503}{.065} & -  \\
    \midrule
    \multirow{7}{*}{\makecell[c]{Human\\ML3D}} & MDM~\cite{tevet2023human} & \et{0.455}{.006} & \et{0.645}{.007} & \et{0.749}{.006} & \et{0.489}{.047} & \et{3.330}{.025} & \etb{9.920}{.083} & \et{2.290}{.070}  \\
    % MDM~\cite{tevet2022human} & - & - & \et{0.611}{.007} & \et{0.544}{.044} & \et{5.566}{.027} & \ets{9.559}{.086} & \etb{2.799}{.072}  \\
    ~ & T2M-GPT~\cite{zhang2023t2m} & \et{0.492}{.003} & \et{0.679}{.002} & \et{0.775}{.002} & \et{0.141}{.005} & \et{3.121}{.009} & \ets{9.761}{.081} & \et{1.831}{.048}  \\
    ~ & ReMoDiffuse~\cite{zhang2023remodiffuse} & \et{0.510}{.005} & \et{0.698}{.006} & \et{0.795}{.004} & \ets{0.103}{.004} & \et{2.974}{.016} & \et{9.018}{.075} & \et{1.795}{.043}  \\
    ~ & FineMoGen~\cite{zhang2023finemogen} & \et{0.504}{.002} & \et{0.690}{.002} & \et{0.784}{.002} & \et{0.151}{.008} & \et{2.998}{.008} & \et{9.263}{.094} & \etb{2.696}{.079}  \\
    ~ & MoMask~\cite{guo2024momask} & \ets{0.521}{.002} & \ets{0.713}{.002} & \ets{0.807}{.002} & \etb{0.045}{.002} & \et{2.958}{.008} & - & \et{1.241}{.040}  \\
    ~ & MotionDiffuse~\cite{zhang2022motiondiffuse} & \et{0.515}{.003} & \et{0.708}{.003} & \et{0.806}{.002} & \et{0.141}{.007} & \ets{2.919}{.008}  & \et{9.485}{.093} & \ets{2.669}{.087}  \\
    \cline{2-9}
    ~ & \textbf{Ours} & \etb{0.524}{.002} & \etb{0.715}{.002} & \etb{0.811}{.001} & \et{0.111}{.005} & \etb{2.879}{.006} & \et{9.527}{.090} & \et{2.604}{.084}  \\
    % \midrule
    \bottomrule
    \end{tabular}
    }
    
    \caption{\textbf{Quantitative evaluation on the test set of HumanML3D.} $\pm$ indicates a 95\% confidence interval. \textbf{Bold} face indicates the best result, while \underline{underscore} refers to the second best.}
    \label{tab:eval_humanml}
        % \vspace{-1.5em}
\end{table*}

\begin{table*}[thb]
    \centering
    \scalebox{0.8}{
    \begin{tabular}{l l c c c c c c c}
    \toprule
    \multirow{2}{*}{Datasets} & \multirow{2}{*}{Methods}  & \multicolumn{3}{c}{R Precision (\%) $\uparrow$} & \multirow{2}{*}{FID$\downarrow$} & \multirow{2}{*}{MM Dist$\downarrow$} & \multirow{2}{*}{Diversity$\rightarrow$} & \multirow{2}{*}{MultiModality$\uparrow$} \\
    \cline{3-5}
    ~ & ~ & Top 1 & Top 2 & Top 3 \\
    \midrule
    \multirow{5}{*}{\makecell[c]{Mixamo}} 
    & MDM~\cite{tevet2023human} & \et{9.338}{.319} & \et{16.168}{.354} & \et{22.091}{.376} & \et{3.537}{.091} & \et{6.135}{.032} & \et{6.981}{.059} & \et{4.505}{.196} \\
    & ReMoDiffuse~\cite{zhang2023remodiffuse} & \et{7.009}{.468} & \et{12.961}{.491} & \et{18.452}{.555} & \ets{3.644}{.061} & \et{6.882}{.076} & \et{2.471}{.188} & \\
    & MoMask~\cite{guo2024momask} & \et{8.484}{.376} & \et{14.141}{.507} & \et{19.790}{.571} & \et{3.669}{.113} & \et{6.175}{.045} & \et{7.024}{.086} & \et{4.017}{.213} \\
    & MotionDiffuse~\cite{zhang2022motiondiffuse} & \et{9.617}{.305} & \et{16.556}{.457} & \et{22.387}{.570} & \et{4.318}{.062} & \et{6.316}{.025} & \et{7.165}{.094} & \et{2.333}{.180} \\
    \cline{2-9}
    & \textbf{Ours} & \etb{10.015}{.338} & \etb{16.875}{.501} & \etb{23.132}{.412} & \et{4.372}{.054} &\et{6.268}{.036} & \et{7.187}{.068} & \et{2.174}{.188} \\
    \bottomrule
\end{tabular}
    % \begin{tabular}{l l c c c c c c c}
    % \toprule
    % \multirow{2}{*}{Datasets} & \multirow{2}{*}{Methods}  & \multicolumn{3}{c}{R Precision (\%) $\uparrow$} & \multirow{2}{*}{FID$\downarrow$} & \multirow{2}{*}{MM Dist$\downarrow$} & \multirow{2}{*}{Diversity$\rightarrow$} & \multirow{2}{*}{MultiModality$\uparrow$}\\

    % \cline{3-5}
    %    ~& ~ & Top 1 & Top 2 & Top 3 \\
    % \midrule
    % \multirow{2}{*}{\makecell[c]{Mixamo}} & MDM~\cite{tevet2023human} & \et{9.338}{.319} & \et{16.168}{.354} & \et{22.091}{.376} & \et{3.537}{.091} & \et{6.135}{.032} & \et{6.981}{.059} & \et{4.505}{.196}  \\
    % ~ & ReMoDiffuse~\cite{zhang2023remodiffuse} & \et{7.009}{.468} & \et{12.961}{.491} & \et{18.452}{.555} & \ets{3.644}{.061} & \et{6.882}{.076} & \et{2.471}{.188} &  \\
    % ~ & MoMask~\cite{guo2024momask} & \et{8.484}{.376} & \et{14.141}{.507} & \et{19.790}{.571} & \et{3.669}{.113} & \et{6.175}{.045} & \et{7.024}{.086} & \et{4.017}{.213}  \\
    % ~ & MotionDiffuse~\cite{zhang2022motiondiffuse} & \et{9.617}{.305} & \et{16.556}{.457} & \et{22.387}{.570} & \et{4.318}{.062} & \et{6.316}{.025} & \et{7.165}{.094} & \et{2.333}{.180} &  
    % \cline{2-9}
    % ~ & \textbf{Ours} & \etb{10.015}{.338} & \etb{16.875}{.501} & \etb{23.132}{.412} & \et{4.372}{.054} &\et{6.268}{.036} & \et{7.187}{.068} & \et{2.174}{.188} \\
    % % \midrule
    % \bottomrule
    % \end{tabular}
    }
    
    \caption{\textbf{Cross-dataset evaluation on Mixamo.} Models are trained on HumanML3D and tested on Mixamo.}
    \label{tab:eval_mixamo}
        % \vspace{-1.5em}
\end{table*}

\begin{table*}[thb]
    \centering
    \scalebox{0.85}{
    \begin{tabular}{l c c c c c c c}
    \toprule
     \multirow{2}{*}{Methods}  & \multicolumn{3}{c}{R Precision (\%) $\uparrow$} & \multirow{2}{*}{FID$\downarrow$} & \multirow{2}{*}{MM Dist$\downarrow$} & \multirow{2}{*}{Diversity$\rightarrow$} & \multirow{2}{*}{MultiModality$\uparrow$}\\

    \cline{2-4}
     ~ & Top 1 & Top 2 & Top 3 \\
    \midrule
     Full body & \et{51.971}{.199} & \et{71.202}{.197} & \et{80.877}{.147} & \et{0.124}{.005} & \et{2.890}{.008} & \et{9.531}{.092} & \et{2.576}{.085}  \\
     Half-body naive & \et{52.090}{.214} & \et{71.167}{.220} & \et{80.776}{.144} & \et{0.144}{.007} & \et{2.894}{.007} & \et{9.506}{.096} & \et{2.585}{.084} \\
     Half-body LLM & \et{52.347}{.231} & \et{71.487}{.245} & \et{81.005}{.133} & \et{0.114}{.005} & \et{2.878}{.007} & \et{9.523}{.090} & \et{2.603}{.084} \\
     \textbf{Ours} & \et{52.351}{.239} & \et{71.450}{.234} & \et{81.069}{.127} & \et{0.111}{.005} & \et{2.879}{.006} & \et{9.527}{.090} & \et{2.604}{.084} \\
    % \midrule
    \bottomrule
    \end{tabular}
    }
    \vspace{-0.3cm}
    \caption{\textbf{Ablation study on different retrieval and composition strategies.}}
    \label{tab:composition}
        % \vspace{-1.5em}
\end{table*}

\begin{table*}[th]
    \centering
    \scalebox{0.85}{
    \begin{tabular}{l c c c c c c c}
    \toprule
     \multirow{2}{*}{Methods}  & \multicolumn{3}{c}{R Precision (\%) $\uparrow$} & \multirow{2}{*}{FID$\downarrow$} & \multirow{2}{*}{MM Dist$\downarrow$} & \multirow{2}{*}{Diversity$\rightarrow$} & \multirow{2}{*}{MultiModality$\uparrow$}\\

    \cline{2-4}
     ~ & Top 1 & Top 2 & Top 3 \\
    \midrule
    startx 8l & \et{49.253}{.367} & \et{68.469}{.339} & \et{78.375}{.279} & \et{0.190}{.008} & \et{3.092}{.009} & \et{9.298}{.081} & \et{2.866}{.096} \\
    startx 8l + ours & \et{49.912}{.293} & \et{69.164}{.268} & \et{78.978}{.248} & \et{0.137}{.006} & \et{3.045}{.008} & \et{9.296}{.092} & \et{2.645}{.084} \\
    \hline
    startx 12l & \et{50.972}{.382} & \et{69.594}{.327} & \et{79.004}{.278} & \et{0.307}{.010} & \et{2.981}{.010} & \et{9.320}{.083} & \et{2.705}{.084} \\
    startx 12l + ours & \et{51.148}{.712} & \et{70.112}{.462} & \et{79.543}{.520} & \et{0.140}{.009} & \et{2.957}{.011} & \et{9.456}{.232} & \et{2.398}{.167} \\
    \hline
    epsilon 8l & \et{50.437}{.218} & \et{69.329}{.259} & \et{78.860}{.243} & \et{0.395}{.009} & \et{3.026}{.011} & \et{9.381}{.094} & \et{2.749}{.085} \\
    epsilon 8l + ours & \et{50.832}{.298} & \et{69.728}{.261} & \et{79.024}{.276} & \et{0.160}{.006} & \et{2.988}{.009} & \et{9.543}{.089} & \et{2.378}{.080} \\
    \hline
    epsilon 12l & \et{51.546}{.326} & \et{70.841}{.291} & \et{80.592}{.165} & \et{0.141}{.007} & \et{2.919}{.008} & \et{9.485}{.093} & \et{2.669}{.087} \\
    epsilon 12l + ours & \et{52.351}{.239} & \et{71.450}{.234} & \et{81.069}{.127} & \et{0.111}{.005} & \et{2.879}{.006} & \et{9.527}{.090} & \et{2.604}{.084} \\
    % \midrule
    \bottomrule
    \end{tabular}
    }
    \vspace{-0.3cm}
    \caption{\textbf{Different base models.} Our method brings consistent improvement for different MotionDiffuse variants.}
    \label{tab:base_model}
        % \vspace{-1.5em}
\end{table*}

\begin{table}[thb]
    \centering
    \scalebox{0.8}{
    \begin{tabular}{l c c c}
    \toprule
     Dataset & Full-body & Half-body & Fine-grained \\
    \midrule
     HumanML3D & 37.8\% & 56.3 \% & 5.9 \\
     Mixamo & 0 & 63.1\% & 36.9\% \\
     User data & 0 & 48.5\% & 51.5\% \\
    % \midrule
    \bottomrule
    \end{tabular}
    }
    \vspace{-0.3cm}
    \caption{\textbf{The percentage of three levels of decomposition and retrieval strategies.} We find real-world data requires more fine-grained decomposition, demonstrating the huge gap between real-world data and standard benchmarks.}
    \label{tab:strategy_frequency}
        % \vspace{-1.5em}
\end{table}

\subsection{Datasets and Metrics}
\paragraph{Datasets.} 
Our evaluation involves two motion datasets: HumanML3D~\cite{guo2022generating} and Mixamo~\cite{mixamo}.
HumanML3D~\cite{guo2022generating} is a standard text-to-motion benchmark.
Derived from HumanAct12 and AMASS datasets, it contains 14,616 motions with 44,970 text descriptions.
To evaluate the methods under out-of-distribution scenario, we curate a testset using Mixamo~\cite{mixamo}, containing over two thousand motion sequences and corresponding text captions.
%

%--------------------------------------------------------------------
\paragraph{Evaluation metrics.} 
We adopt five standard metrics following previous works~\cite{guo2022generating,zhang2022motiondiffuse}:

1. R-Precision: Assesses text-motion alignment by calculating the probability of matching the correct text description within the top $k$ candidates ($k$ = 1, 2, 3).

2. Frechet Inception Distance (FID): Measures generation quality by computing the distance between real and generated motion features.

3. Diversity: Quantifies the overall variety among generated motion sequences.

4. Multimodality: Measures the variation in motion sequences generated from a single text prompt.

5. Multi-Modal Distance (MM Dist): Calculates the average Euclidean distance between motion and text features.

%--------------------------------------------------------------------
\subsection{Implementation Details}
We employ GPT-4o (2024-05-01-preview) as the LLM in this work.
For measuring the semantic similarity, we use the frozen text encoder in the CLIP ViT-B/32~\cite{radford2021learning}.
To construct the retrieval database, we simply use all the training data as the entries.
For the diffusion model, we re-train MotionDiffuse~\cite{zhang2022motiondiffuse}, consisting of 12 transformer layers, which has better performance than the version in the original paper.
During inference, we apply DDIM sampling with a total step of 50.
When not specified otherwise, we choose the retrieval length coefficient $\lambda=0.05$, the retrieval thresholds $\tau_{half}=0.96, \tau_{fine}=0.96$, and the diffusion starting time $t_0=0.96$.

\begin{figure*}
    \centering
    \includegraphics[width=0.95\linewidth]{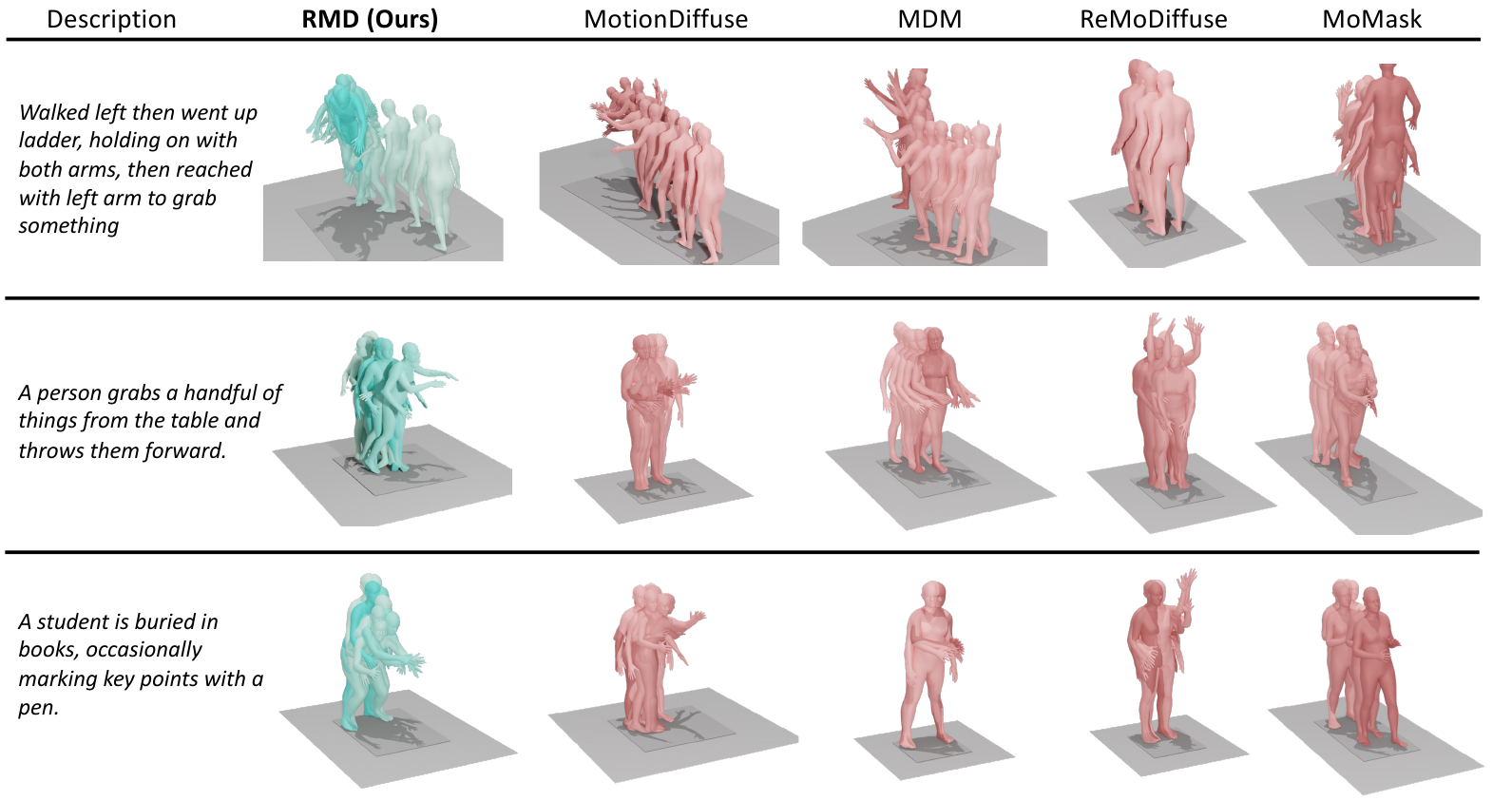}
    \vspace{-0.2cm}
    \caption{\textbf{Qualitative comparison between our method and previous methods.} Our method achieves the best text alignment. }
    \label{fig:comparison}
\end{figure*}

\begin{figure*}
    \centering
    \includegraphics[width=0.95\linewidth]{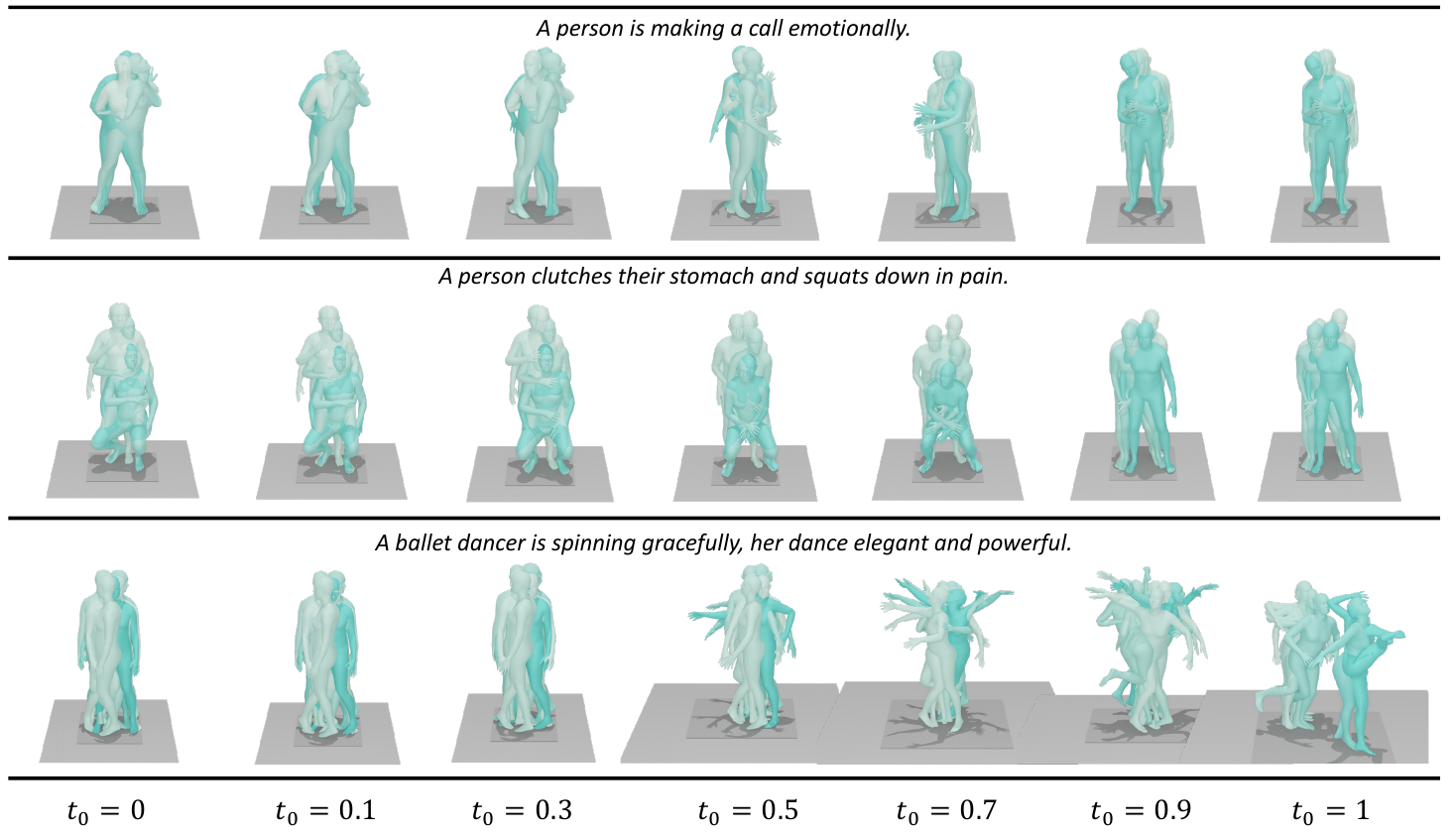}
    \vspace{-0.2cm}
    \caption{\textbf{Generated motions with various $t_0$.} $t_0=0$ is the guided motion. $t_0=1$ means starting from pure noise and is equivalent to MotionDiffuse. Since we use the same random seed for all samples here. $t_0$ in between can be seen as an interpolation between the guided motion and the pure diffusion generation. In the first two rows, the guided motion has obvious artifacts, while MotionDiffuse fails to understand the prompt. Yet a $t_0$ in the middle incorporates the semantic information from the guided motion while free from artifacts. In the third row, the guided motion retrieves a spinning motion without dancing, while MotionDiffuse generates dance without spinning. A proper $t_0$ could combine these information to produce better results.}
    \label{fig:t0}
\end{figure*}

%--------------------------------------------------------------------
\subsection{Comparison}
We present quantitative and qualitative comparisons between our proposed method and several state-of-the-art approaches on the HumanML3D dataset. 
As shown in Table \ref{tab:eval_humanml}, our approach achieves the best R Precision scores and MM Dist.
Compared to our base model, MotionDiffuse, our method shows superior performance across all metrics except a slight drop in MultiModality, demonstrating that our method brings instant improvement to the existing diffusion model without training.
Figure \ref{fig:comparison} provides a visual comparison, highlighting the qualitative differences between our method and previous approaches.
Our generated motions appear more realistic and align better with the text descriptions.

\begin{figure}[H]
    \centering
    \includegraphics[width=0.8\linewidth]{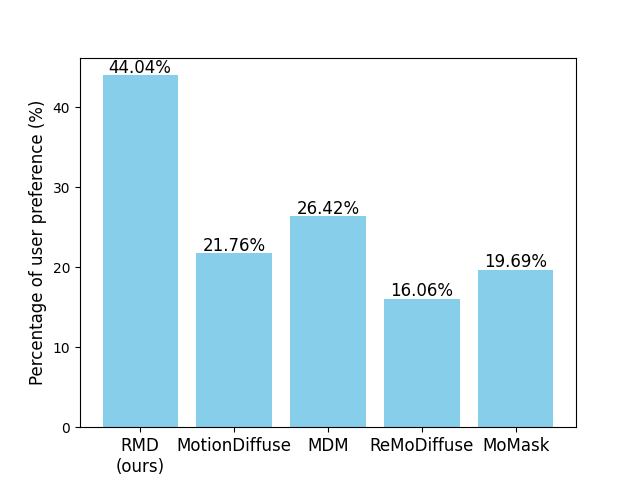}
    \vspace{-0.3cm}
    \caption{\textbf{User study on OOD data.} Our method outperforms others by a significant margin.}
    \vspace{-0.3cm}
    \label{fig:user_study}
\end{figure}

%--------------------------------------------------------------------
\paragraph{Cross-dataset evaluation.}
HumanML3D has recently become the most popular text-to-motion benchmark, and recent methods have achieved good quantitative performance on it.
However, to better evaluate the generalization ability, we conduct a cross-dataset evaluation, where all the models are trained or retrieved using HumanML3D, and tested on Mixamo.
We retarget all motions in Mixamo to the skeleton of HumanML3D, and convert them to the pose representation we use.
We only keep the motion sequences that are between 2 seconds and 10 seconds.
Out of these 1294 sequences, we equally divide them into two splits, one for training the evaluator~\cite{guo2022generating} and the other for evaluation.
The results are presented in Table~\ref{tab:eval_mixamo}.
Our method shows better performance regarding R-precisions, denoting better semantic distinctiveness.
Although our FID and MultiModality are slightly worse than others, \name still maintains the performance of MotionDiffuse, the base model we use.

%--------------------------------------------------------------------
\paragraph{User study. }
To further evaluate the generalization ability of our method, we conduct a user study on OOD text prompts.
To construct the test set, we asked a 3D animator to collect some short videos containing human movement on Youtube, and write down a text description for each, resulting in 33 text prompts.
We run our method and all the competing methods on these prompts and shuffle the order.
We conducted a survey asking users which motion output they prefer regarding text alignment and motion naturalness.
For each survey question, users are allowed to make multiple selections or none at all.
We collect feedback from 16 users and calculate the preference percentages of each method in Figure~\ref{fig:user_study}.
Our method significantly outperforms all the other methods, demonstrating a strong ability to generalize to OOD data.
Interestingly, as two early methods, MDM and MotionDiffuse outperformed the recent state-of-the-art MoMask.
Although ReMoDiffuse is also retrieval-based, visualizations suggest that injecting retrieval results into the training process may lead to overfitting.
%
% Refer to supplementary materials for more details.

%--------------------------------------------------------------------
\subsection{Analysis}

\begin{figure}[H]
    \centering
    \includegraphics[width=0.7\linewidth]{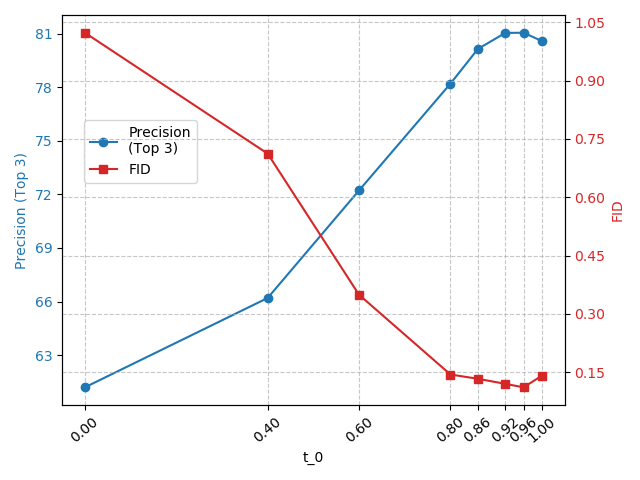}
    \vspace{-0.5cm}
    \caption{The performance on HumanML3D of different $t_0$}
    \label{fig:t0_humanml}
\end{figure}

\paragraph{The impact of $t_0$.}
Similar to SDEdit~\cite{meng2021sdedit},
the choice of $t_0$ is an important factor in our method.
In Fig.~\ref{fig:t0}, we visualize how the generated motion changes with various $t_0$.
When using larger $t_0$, the output is closer to the pure random diffusion sampling.
Otherwise, it is closer to the guided motion formed by our retrieval and composition process.
We also evaluate the impact of $t_0$ on HumanML3D in Fig.~\ref{fig:t0_humanml}.
We find the performance reaches a peak when $t_0=0.96$ and choose this value for our main comparisons.
%
%--------------------------------------------------------------------

\paragraph{Ablation study of retrieval and composition strategies.}
To validate our design's effectiveness, we ablate on different retrieval and composition strategies in Table~\ref{tab:composition}.
Here, ``Full body'' denotes that only full body retrieval is used. 
``Half-body'' denotes full-body and half-body decomposition, which is used without the fine-grained decomposition.
``Half-body naive'' represents the naive retrieval strategy, while ``Half-body LLM'' denotes the LLM retrieval strategy instead.
Furthermore, we calculate the percentage of each level of retrieval used in different datasets and show the results in Table~\ref{tab:strategy_frequency}.
We can see that HumanML3D test set has a high proportion of full-body hits because it comes from the same distribution as our database, \ie the training set of HumanML3D.
As the dataset deviates from the distribution of HumanML3D, the proportion of more fine-grained retrieval increases.
We notice the data used for the user study needs the most fined-grained retrieval, demonstrating a huge gap between real-world data and standard benchmarks.

%--------------------------------------------------------------------
\paragraph{Different base models.}
We experiment with different architectures of MotionDiffuse to validate the robustness of our method in Table~\ref{tab:base_model}.
``startx'' denotes that the output of the network is $\motion(0)$  and ``epsilon'' denotes the network predicts the noise residule.
``8l'' and ``12l'' represent 8 layers and 12 layers of transformer layers, respectively.
It can be seen that our model bring consistent performance gain to all variants of the diffusion model.

%-------------user study---------------------

% \begin{table}[thb]
%     \centering
%     \resizebox{0.95\columnwidth}{!}{%
%     \begin{tabular}{l|ccc} \hline
%     Method & MoMask & MotionDiffuse & Ours \\ \hline
%     User preference (\%) & 27 & 9 & 31 \\ \hline
%     \end{tabular}%
%     }
%     \caption{Percentage of user preferences for each method.}
%     \label{tab:user_study}
%     \end{table}

\section{Conclusion and Future Work}
In conclusion, we propose \name, a simple, training-free baseline towards more general motion generation.
% In conclusion, we propose \name, a simple, training-free baseline to motion generation that effectively addresses out-of-distribution scenarios.
%
By combining a retrieval-augmented framework with a pre-trained motion diffusion model, \name outperforms existing methods in both in-domain and cross-domain settings.
Furthermore, user studies validate \name's ability to generate natural and semantically accurate motions, establishing a new benchmark for future motion generation research.

\paragraph{Limitations and future work.}
Currently, $t_0$ in our method is either fixed or specified by users.
While this approach enhances user controllability,, automating $t_0$ selection could improve convenience and potentially yield better performance on standard benchmarks.
%
% Meanwhile, while our method brings a performance boost on different versions of MotionDiffuse., we didn't notice a significant boost when plugging it into MDM [??? should we write this].
% Although \name outperforms other methods, quantitative and qualitative results reveal significant room for improvement in general motion generation for real-world applications.
% \clearpage
{
    \small
    \bibliographystyle{ieeenat_fullname}
    \bibliography{main}
}

\clearpage
\setcounter{page}{1}
\maketitlesupplementary
\begin{table}[b]
    \centering
    \scalebox{0.9}{
    \begin{tabular}{l c c c c c}
    \toprule
     \multirow{2}{*}{Methods}  & \multicolumn{3}{c}{R Precision (\%) $\uparrow$} & \multirow{2}{*}{FID$\downarrow$} & \multirow{2}{*}{MM Dist$\downarrow$} \\

    \cline{2-4}
     ~ & Top 1 & Top 2 & Top 3 \\
    \midrule
     SDS~\cite{poole2022dreamfusion} & 48.909 & 66.653 & 77.018 & 0.136 & 3.160 \\
     DDS~\cite{hertz2023delta} & 40.883 & 59.317 & 69.674 & 1.263 & 3.753 \\ % 8.438 & 2.241
     DNO~\cite{karunratanakul2024optimizing} & 49.570 & 69.449 & 79.346 & 0.293 & 3.130 \\
     % \midrule
     % STMC~\cite{petrovich2024multi} & 23.312 & 35.979 & 44.524 & 5.712 & 5.588 \\ % & 7.259 & 1.528 \\
     \midrule
     \textbf{Ours} & 52.351 & 71.450 & 81.069 & 0.111 & 2.879 \\
    % \midrule
    \bottomrule
    \end{tabular}
    }
    \caption{Comparison with other baseline methods.}
    \label{tab:inversion}
        % \vspace{-1.5em}
\end{table}

\begin{table*}[th]
    \centering
    \scalebox{0.92}{
    \begin{tabular}{l c c c c c c c}
    \toprule

      \multirow{2}{*}{Methods}  & \multicolumn{3}{c}{R Precision$\uparrow$} & \multirow{2}{*}{FID$\downarrow$} & \multirow{2}{*}{MM Dist$\downarrow$} & \multirow{2}{*}{Diversity$\rightarrow$} & \multirow{2}{*}{MultiModality$\uparrow$}\\

    \cline{2-4}
     ~ & Top 1 & Top 2 & Top 3 \\
    \midrule
    MotionDiffuse~\cite{zhang2022motiondiffuse} & \et{44.512}{.315} & \et{63.230}{.276} & \et{73.414}{.244} & \et{0.974}{.013} & \et{3.329}{.010}  & \et{8.674}{.095} & \etb{2.539}{.092}  \\
    \midrule
     Ours ($t_0=0$) & \et{34.996}{.220} & \et{51.360}{.189} & \et{61.184}{.206} & \et{1.022}{.001} & \et{4.179}{.008} & \et{8.739}{.087} & \et{0.014}{.000}  \\
     Ours ($t_0=0.4$) & \et{38.521}{.259} & \et{55.487}{.215} & \et{65.187}{.273} & \et{0.989}{.005} & \et{3.892}{.008} & \et{8.980}{.080} & \et{0.789}{.030}  \\
     Ours ($t_0=0.5$) & \et{40.548}{.254} & \et{57.829}{.250} & \et{67.527}{.286} & \et{0.852}{.007} & \et{3.729}{.010} & \et{8.998}{.075} & \et{1.158}{.036}  \\
     Ours ($t_0=0.6$) & \et{42.766}{.255} & \et{60.344}{.281} & \et{70.132}{.309} & \ets{0.552}{.007} & \et{3.530}{.010} & \ets{9.056}{.082} & \et{1.517}{.039}  \\
     Ours ($t_0=0.7$) & \et{44.939}{.343} & \et{62.925}{.351} & \et{72.756}{.273} & \etb{0.432}{.007} & \et{3.337}{.010} & \etb{9.063}{.085} & \et{1.814}{.046}  \\
     Ours ($t_0=0.76$) & \et{45.944}{.333} & \et{64.414}{.286} & \et{74.177}{.242} & \et{0.507}{.009} & \et{3.234}{.009} & \et{9.040}{.078} & \et{2.010}{.063}  \\
     Ours ($t_0=0.8$) & \et{46.281}{.239} & \et{65.097}{.288} & \et{75.120}{.273} & \et{0.585}{.013} & \et{3.191}{.009} & \et{9.005}{.080} & \et{2.143}{.068}  \\
     Ours ($t_0=0.84$) & \etb{46.511}{.251} & \etb{65.487}{.207} & \ets{75.546}{.182} & \et{0.693}{.015} & \etb{3.170}{.008} & \et{8.948}{.080} & \et{2.314}{.076}  \\
     Ours ($t_0=0.88$) & \ets{46.266}{.209} & \ets{65.254}{.286} & \etb{75.548}{.232} & \et{0.718}{.013} & \ets{3.175}{.007} & \et{8.938}{.083} & \et{2.375}{.080}  \\
     Ours ($t_0=0.92$) & \et{46.028}{.251} & \et{64.857}{.289} & \et{75.104}{.195} & \et{0.759}{.013} & \et{3.207}{.008} & \et{8.896}{.086} & \et{2.456}{.082}  \\
     Ours ($t_0=0.96$) & \et{45.666}{.298} & \et{64.236}{.313} & \et{74.499}{.187} & \et{0.821}{.013} & \et{3.249}{.008} & \et{8.842}{.079} & \ets{2.495}{.084}  \\
    \bottomrule
    \end{tabular}
    }
    
    \caption{\textbf{Performance evaluation on HumanML3D test set with misaligned retrieval base and training set.} The diffusion model is trained on half of the training set, and the retrieval is conducted on the complete training set. This experiment highlights the effectiveness of our approach in scenarios where the retrieval database exceeds the training set.}
    \label{tab:subset_exp}
\end{table*}

\begin{table*}[th]
    \centering
    \scalebox{0.92}{
    \begin{tabular}{l c c c c c c c}
    \toprule

      \multirow{2}{*}{Methods}  & \multicolumn{3}{c}{R Precision$\uparrow$} & \multirow{2}{*}{FID$\downarrow$} & \multirow{2}{*}{MM Dist$\downarrow$} & \multirow{2}{*}{Diversity$\rightarrow$} & \multirow{2}{*}{MultiModality$\uparrow$}\\

    \cline{2-4}
     ~ & Top 1 & Top 2 & Top 3 \\
    %\midrule
     %   \textbf{Real motions} & \et{0.511}{.003} & \et{0.703}{.003} & \et{0.797}{.002} & \et{0.002}{.000} & \et{2.974}{.008} & \et{9.503}{.065} & -  \\
    \midrule
    MotionDiffuse~\cite{zhang2022motiondiffuse} & \et{41.895}{.781} & \et{63.696}{.558} & \et{75.693}{.592} & \et{0.328}{.014} & \et{2.994}{.016}  & \et{10.798}{.103} & \etb{1.557}{.098}  \\
    \midrule
     \textbf{Ours} & \etb{43.336}{.756} & \etb{65.211}{.529} & \etb{77.629}{.416} & \etb{0.320}{.019} & \etb{2.863}{.015} & \etb{10.879}{.110} & \et{1.364}{.100}  \\
    % \midrule
    \bottomrule
    \end{tabular}
    }
    
    \caption{\textbf{Quantitative evaluation on the test set of KIT-ML.} Our model brings significant improvement to the base diffusion model.}
    \label{tab:kit}
        % \vspace{-1.5em}
\end{table*}

\begin{table*}[b]
\centering
\begin{tabular}{p{0.13\textwidth}|p{0.87\textwidth}}
\hline
\multicolumn{1}{c|}{\textbf{Task}} & \multicolumn{1}{c}{\textbf{Prompt}} \\ 
\hline
\makecell[l]{Half-body \\ decomposition} & \makecell[l]{The following sentence describes a human motion.\\ According to it, write two sentences for the upper body motion and the lower body motion for that motion.\\ The sentences should be brief. The answer should be two lines (no empty lines), \\the first line for the upper body motion and the second line for the lower body motion.\\ Two lines are separated by a newline character. A subject is not needed in the sentence.\\ There should also not be a label such as "Upper body motion:" in the sentences.\\ Motion: [description] } \\ 
\hline
\makecell[l]{Fine-grained \\ decomposition} & \makecell[l]{The following sentence describes a human motion.\\ According to it, decompose into seven sentences that describe the motions of seven body parts: \\ head, spine, left upper limb, right upper limb, left lower limb, right lower limb, \\ and the overall displacement/orientation(trajectory).\\ The sentences should be brief.\\ The answer should be seven lines separated by newline characters, following the order above.\\ A subject such as "the person" is not needed in the sentence.\\ There must not be any label such as "Head:" or "Spine:" in the sentences.\\ Motion: [description]} \\ 
\hline
Retrieval agent & \makecell[l]{[retrieved motion prompts] \\ Which sentence above best describes the [part] motion of this sentence [original motion prompt]?\\ Briefly explain your reasoning and put the answer sentence index without any symbol in the last line.} \\ 
\hline
\end{tabular}
\caption{LLM prompts used for motion decomposition and retrieval.}
\label{tab:prompts}
\end{table*}

\begin{table*}[]
\centering

\begin{subtable}{\textwidth}
\centering
\begin{tabular}{p{0.15\textwidth}|p{0.75\textwidth}}
\hline
Original   & A person walks anti clockwise and holds left hand above head                                        \\ \hline
Upper body & Left hand remains raised above the head.                  \\ 
Lower body & Feet move in an anti-clockwise spiral pattern.     \\ \hline
Head       & The head remains stable as the body moves.                           \\ 
Torso      & The spine maintains an upright posture, slightly twisting with the circular motion.                     \\ 
Left arm   & The left upper limb is raised above the head.  \\ 
Right arm  & The right upper limb swings naturally by the side. \\ 
Trajectory & The overall trajectory follows an anti-clockwise spiral path.                       \\ \hline
\end{tabular}
\end{subtable}
\vspace{0.22cm}

\begin{subtable}{\textwidth}
\centering
\begin{tabular}{p{0.15\textwidth}|p{0.75\textwidth}}
\hline
Original   & A person jumps sideways to the left                                        \\ \hline
Upper body & Arms swing naturally to maintain balance during the jump.                  \\ 
Lower body & Legs push off the ground and land to the left, shifting body sideways.     \\ \hline
Head       & Head moves in a sideways trajectory to the left.                           \\ 
Torso      & Spine remains aligned and moves laterally to the left.                     \\ 
Left arm   & Left upper limb moves in coordination with the sideways jump to the left.  \\ 
Right arm  & Right upper limb moves in coordination with the sideways jump to the left. \\ 
Trajectory & Overall displacement is a sideways jump to the left.                       \\ \hline
\end{tabular}
\end{subtable}
\vspace{0.22cm}

\begin{subtable}{\textwidth}
\centering
\begin{tabular}{p{0.15\textwidth}|p{0.75\textwidth}}
\hline
Original   & The person was walking forward on a balance beam.                                        \\ \hline
Upper body & Arms extended for balance.                  \\ 
Lower body & Feet placed heel-to-toe along the beam.     \\ \hline
Head       & Head maintains a forward tilt for balance.                           \\ 
Torso      & Spine remains straight to assist in balance.                     \\ 
Left arm   & Left upper limb held steady or slightly out to the side for balance.  \\ 
Right arm  & Right upper limb mirrors the left for balance. \\ 
Trajectory & Overall movement progresses forward along a linear path on the balance beam.                       \\ \hline
\end{tabular}
\end{subtable}
\vspace{0.22cm}

\begin{subtable}{\textwidth}
\centering
\begin{tabular}{p{0.15\textwidth}|p{0.75\textwidth}}
\hline
Original   & A person who seems to be warming up their left leg                                        \\ \hline
Upper body & Lifts arms for balance while shifting weight.                  \\ 
Lower body & Raises one knee high and swings the other leg behind.     \\ \hline
Head       & Head remains steady and upright.                           \\ 
Torso      & Spine maintains an upright posture with slight adjustments for balance.                     \\ 
Left arm   & Left upper limb stays relaxed at the side or extends slightly for balance.  \\ 
Right arm  &Right upper limb is held in a position to aid balance. \\ 
Trajectory & Overall balance on left leg, with the body adjusting for coordination and stability.                       \\ \hline
\end{tabular}
\end{subtable}
\vspace{0.22cm}

\begin{subtable}{\textwidth}
\centering
\begin{tabular}{p{0.15\textwidth}|p{0.75\textwidth}}
\hline
Original   & A programmer is typing on a laptop, occasionally scratching his hair.                                        \\ \hline
Upper body & Arms are reaching forward to type, with one hand intermittently moving to scratch the head.            \\ 
Lower body & Legs remain still, providing a stable seated posture.     \\ \hline
Head       & Head is tilted slightly downward toward the screen.                           \\ 
Torso      & Spine is leaned slightly forward, maintaining an ergonomic posture.                   \\ 
Left arm   & Left upper limb is raised to scratch the hair intermittently.  \\ 
Right arm  & Right upper limb is extended forward, fingers tapping on the keyboard. \\ 
Trajectory & Remains seated on the chair with minimal movement.                 \\ \hline
\end{tabular}
\end{subtable}
\vspace{0.22cm}

\begin{subtable}{\textwidth}
\centering
\begin{tabular}{p{0.15\textwidth}|p{0.75\textwidth}}
\hline
Original   & A person holds a cigarette near the mouth with their left hand and lights it with the right hand.                            \\ \hline
Upper body & Left hand brings the cigarette close to the mouth while right hand flicks the lighter.          \\ 
Lower body & Feet maintain a balanced stance during the action.   \\ \hline
Head       & Head remains steady.                 \\ 
Torso      & Spine maintains an upright posture.            \\ 
Left arm   & Left upper limb moves to bring the cigarette near the mouth.  \\ 
Right arm  & Right upper limb holds a lighter and moves to ignite the cigarette. \\ 
Trajectory & Overall body maintains its position without displacement.     \\ \hline
\end{tabular}
\end{subtable}
% 3850 0002 9769 3880 user/1 user/21
\caption{Examples of motion decomposition.}
\label{example_decomposition}
\end{table*}
\begin{table*}[]
\centering

\begin{subtable}{\textwidth}
\centering
\begin{tabular}{p{0.11\textwidth}|p{0.88\textwidth}}
\hline
Original   & A person points forward with their right hand. \\ \hline
Upper body & 1. Extends right arm forward with the hand pointing or pressing. \\ 
decomposition & 2. Extends right arm forward.  \\
 & 3. Extends right arm forward. Moves right hand towards a point. \\
 & 4. Extends the right arm forward. \\
 & 5. Right arm extends forward. \\
\hline
Retrieved       & \textbf{1. Extends right arm forward to point or touch.}  \\ 
descriptions      & 2. Extends right arm forward swiftly. \\ 
   & 3. Right arm extends forward and then lowers. \\ 
  & \fbox{4. Raises right arm quickly.} \\ 
 & 5. Right arm extended outward.  \\ \hline
\end{tabular}
\end{subtable}
\vspace{0.5cm}
% 2049

\begin{subtable}{\textwidth}
\centering
\begin{tabular}{p{0.11\textwidth}|p{0.88\textwidth}}
\hline
Original   & This man bends down to pick up a box or ball, and then puts it on a shelf, which is on his chest level. \\ \hline
Upper body & 1. Arms extend forward to pick up an item and place it at waist height. \\ 
decomposition & 2. Bends at the waist, extends arms forward to pick up an object, then raises it to chest level.  \\
 & 3. Bending forward at the waist, arms reach down and then extend forward at waist level. \\
 & 4. Bends forward at the waist, reaches down, and raises arms to chest level. \\
 & 5. Bends forward at the waist, reaches down and then places at chest level. \\
\hline
Retrieved       & 1. Arms hang by the sides, then extend forward to pick up an item and return it.  \\ 
descriptions      & \textbf{2. Bends forward and grabs object with both hands, lifts it to chest level and rotates arms outwards.} \\ 
   & 3. Bending forward at the waist while moving backward. \\ 
  & 4. Bends at the waist and raises arms in front. \\ 
 & \fbox{5. Bends the upper body forward from the waist and then rises back up.} \\ \hline
\end{tabular}
\end{subtable}
\vspace{0.5cm}
% 2055

\begin{subtable}{\textwidth}
\centering
\begin{tabular}{p{0.11\textwidth}|p{0.88\textwidth}}
\hline
Original   & A man runs to the right then runs to the left then back to the middle.                            \\ \hline
Lower body & 1. Legs push off the ground alternately in a rhythmic jog.          \\ 
decomposition & 2. Legs engage in a steady, rhythmic stride shifting direction between left, right, and center.   \\
 & 3. Legs alternate between stepping forward in a jogging pattern. \\
 & 4. Legs move in a rhythmic, alternating pattern to the left and right. \\
 & 5. Legs alternate in a left to right jogging pattern, then return to the center. \\
\hline
Retrieved       & 1. Legs push off the ground diagonally before transitioning into a run.        \\ 
descriptions      & 2. Legs move in a rhythmic stride, shifting left, right, and back to starting position.        \\ 
   & 3. Legs moving in a repetitive jogging pattern, alternating between sides.  \\ 
  & \fbox{4. Legs move in a rhythmic, alternating pattern.} \\ 
 & \textbf{5. Legs move in a jogging pattern, changing direction from right to left, and back to the original spot.}    \\ \hline
\end{tabular}
\end{subtable}
\vspace{0.5cm}
% 0019

\begin{subtable}{\textwidth}
\centering
\begin{tabular}{p{0.11\textwidth}|p{0.88\textwidth}}
\hline
Original   & A person walks in a counterclockwise circle.                       \\ \hline
Lower body & 1. Feet move sequentially in a circular path, taking eight steps to complete the rotation.          \\ 
decomposition & 2. Legs step evenly in a circular path.  \\
 & 3. Feet step sequentially in a circular path. \\
 & 4. Feet alternate in a steady rhythm, moving in a counterclockwise circular path. \\
 & 5. Feet move in a continuous circular path. \\
\hline
Retrieved       & 1. Feet step alternately to the right and then continue stepping to form a counterclockwise circular path.     \\ 
descriptions      & 2. Legs step in a circular path.      \\ 
   & \fbox{3. Feet step alternately in a circular path.}  \\ 
  & \textbf{4. Feet move in a counterclockwise circular path.} \\ 
 & 5.Feet move in a circular path in a clockwise direction.    \\ \hline
\end{tabular}
\end{subtable}
% 0021

\caption{\textbf{Examples of retrieval agent.} \textbf{Bold} face indicates the entry selected by the LLM retrieval agent. The \fbox{box} refers to the one selected by the naive retrieval strategy, which directly chooses the one with the highest similarity score. }
\label{tab:example_retrieval}
\end{table*}

\section{Training and Retrieval Set Mismatch}

In real-world applications, the training set for the diffusion model and the retrieval database often do not align perfectly. Typically, the retrieval database grows progressively as more motions are collected, while the diffusion model cannot always be updated in a timely manner. To simulate such scenarios, we trained MotionDiffuse~\cite{zhang2022motiondiffuse} using only half of the training set from HumanML3D~\cite{guo2022generating} and conducted the retrieval on the complete training set.

Our evaluation, presented in Table~\ref{tab:subset_exp}, assesses the performance of our model on the test set of HumanML3D with varying choices of $t_0$. Notably, our model achieves the best performance when $t_0$ is set around $0.8$, significantly outperforming MotionDiffuse. This experiment demonstrates a more pronounced performance boost compared to the full training set experiment on HumanML3D (Table~\ref{tab:eval_humanml}), underscoring the effectiveness of our approach in scenarios where the retrieval database exceeds the training set.

Furthermore, the optimal $t_0$ in this experiment setting is lower than that observed in the full HumanML3D experiment (as seen in Figure~\ref{fig:t0_humanml}). This observation suggests that retrieval becomes more critical when the retrieval database is larger than the training set for the diffusion model.

\section{Evaluation on KIT-ML}
We further assess the efficacy of our approach on the KIT-ML~\cite{plappert2016kit} dataset, which comprises 3,911 motion sequences and 6,363 corresponding text descriptions.
We employ MotionDiffuse~\cite{zhang2022motiondiffuse} as our baseline, which is trained using a 4-layer transformer architecture. 
Subsequently, we integrate our method with this base diffusion model, setting $t_0 = 0.9$.
The results, as in Table~\ref{tab:kit}, demonstrate a substantial enhancement in performance when utilizing our method, without the need for additional training.

\section{Other Baseline Methods}
In addition to SDEdit, we explored several other methods that leverage diffusion priors to generate motions based on guided inputs.
The results are summarized in Table~\ref{tab:inversion}, where our method demonstrates a significant performance advantage over all other baselines.
\par
%%% SDS
\textbf{SDS} (Score Distillaion Sampling)~\cite{poole2022dreamfusion} is an important tool that optimizes the image using the pre-trained diffusion prior.
We use MotionDiffuse as the diffusion prior and optimize the guided motion to obtain the final motion.
%%% DDS
\par
\textbf{DDS} (Delta Denoising Score)~\cite{hertz2023delta}  is tailored for image editing, leveraging the rich prior of image diffusion models.
Unlike SDS, DDS incorporates the gradient of the initial reference image to guide the optimization process.
In our application, we utilized the guided motion as the reference sample to provide directional cues.
The optimization objective is formulated as:
\begin{equation}
\nabla_{\Theta}\mathcal{L}_{\text{DDS}} =  \epsilon^{\omega}_{\phi} ( \Theta_{\mathbf{t}}, y, t ) - \epsilon^{\omega}_{\phi} ( \hat{\Theta}_{\mathbf{t}}, \hat{y}, t) ,
\end{equation}
where $\Theta$ is the motion to be optimized, $\Theta_{\mathbf{t}}$ is the noised $\Theta$, $\hat{\Theta}_{\mathbf{t}}$ is the noised version of the guided motion, $y$ is the motion prompt, and $\hat{y}$ is the description of the retrieved sample. Note that $\Theta_{\mathbf{t}}$ and $\hat{\Theta}_{\mathbf{t}}$ share the same sample noise.
When the guided motion consists of multiple parts, we independently feed each part into the denoising network and subsequently combine the results using part-wise masks.
\par
%%% DNO
\textbf{DNO}~\cite{karunratanakul2024optimizing} achieves DDIM inversion by optimizing the initial noise through gradient descent.
We set the guided motion as the generation target and optimize the initial noise.
Then, we start from the optimized initial noise to run the whole DDIM.
% Besides, we compare our method with STMC~\cite{petrovich2024multi}, which utilizes a pre-trained motion diffusion network to generate motions composed of multiple spatial or temporal sub-motions.

% \section{Pseudo-code}
\section{LLM Agent Examples}
Table~\ref{example_decomposition} shows examples of our motion decomposition agent.
Table~\ref{tab:example_retrieval} shows examples of our retrieval agent.
In constrast to the naive retrieval strategy, which directly chooses the retrieval entry with the highest similarity score, our method achieves better semantic accuracy.
Our prompts used for both agents are presented in Table~\ref{tab:prompts}.

%%%%%%%%%%%%%%%%%%%%%%%%%%%%%%%%%

\end{document}